\documentclass[a4paper,conference]{IEEEtran}

\usepackage{epsfig}
\usepackage{graphicx}
\usepackage{amsmath}
\usepackage{amssymb}
\usepackage{booktabs}
\usepackage{adjustbox}
\usepackage{enumitem}
\usepackage{multirow}
\usepackage{marvosym}
\usepackage{caption}
\usepackage[pagebackref=true,breaklinks=true,colorlinks,bookmarks=false]{hyperref}

\begin{document}

\title{Unsupervised Learning of Facial Landmarks based on Inter-Intra Subject Consistencies}

\author{\IEEEauthorblockN{Weijian Li\IEEEauthorrefmark{1},
Haofu Liao\IEEEauthorrefmark{1},
Shun Miao\IEEEauthorrefmark{2},
Le Lu\IEEEauthorrefmark{2} and
Jiebo Luo\IEEEauthorrefmark{1}}
\IEEEauthorblockA{\IEEEauthorrefmark{1}Department of Computer Science, 
University of Rochester, Rochester, NY, USA}
\IEEEauthorblockA{\IEEEauthorrefmark{2}PAII. Inc., Bethesda, MD, USA}
\IEEEauthorblockA{Email:\IEEEauthorrefmark{1}\{wli69, hliao6, jluo@cs.rochester.edu\}, \IEEEauthorrefmark{2}\{shwinmiao, tiger.lelu@gmail.com\}}}

\maketitle

%%%%%%%%%%%%%%%%%%%%%%%%%%%%%%
\begin{abstract}
We present a novel unsupervised learning approach to image landmark discovery by incorporating the inter-subject landmark consistencies on facial images. This is achieved via an inter-subject mapping module that transforms original subject landmarks based on an auxiliary subject-related structure. To recover from the transformed images back to the original subject, the landmark detector is forced to learn spatial locations that contain the consistent semantic meanings both for the paired intra-subject images and between the paired inter-subject images. Our proposed method is extensively evaluated on two public facial image datasets (MAFL, AFLW) with various settings. Experimental results indicate that our method can extract the consistent landmarks for both datasets and achieve better performances compared to the previous state-of-the-art methods quantitatively and qualitatively. 
\end{abstract}
%%%%%%%%%%%%%%%%%%%%%%%%%%%%%%

\IEEEpeerreviewmaketitle

%%%%%%%%%%%%%%%%%%%%%%%%%%%%%%
\section{Introduction}
Facial landmark localization aims to detect a set of semantic keypoints on the given objects from images, such as the eyes, nose, and ears of human faces. It has been an essential process to assist many high-level computer vision tasks~\cite{bulat2018super,siarohin2019animating}. Traditional fully supervised approach relies on a set of annotated landmark locations that are labeled by human experts. These landmarks are subsequently used to train a supervised model before it can be applied to unseen images. Although many efforts have been made in this direction and promising results have been achieved~\cite{xiong2015global,wang2019adaptive,feng2018wing,wu2018look,li2020structured,hassaballah2019facial,hassaballah2011automatic}, the challenge of supervised models remains that a large amount of human labeling efforts are required to have desirable performance, which is expensive and the annotation processing is subjective. % could be biased by annotator's different landmark understandings.

Another recent approach follows the unsupervised learning strategy to extract keypoints with self-supervision~\cite{thewlis2017unsupervised,jakab2018unsupervised,sanchez2019object,thewlis2019unsupervised}. Many of the existing methods propose to apply a group of random transformations, such as rotations and translations, on the original image to generate the transformed and paired images. Machine learning models are trained to predict landmark locations based on the fact and constraint that the paired landmarks should follow the same transformation.

Despite the popularity and success, training landmark detectors with only paired images from the same subject images may be insufficient to discover the inter-subject consistency among different subjects. The trained detector may be biased to learn landmark locations that are meaningful for the transformation within the same-subject pairs, but make different predictions on the same landmark across different subjects. 

To this end, we propose a novel unsupervised learning method for image landmark discovery via exploring and integrating on the inter-subject consistency. Our method follows the standard equivariance approach by using image reconstruction as supervision cues, added with injecting a subject mapping module between the image encoder and decoder to ensure the inter-subject landmark semantics. Specifically, (1) our model first extracts the feature maps from the input image, then computes a landmark heatmap from an auxiliary subject image as the structural guidance. (2) We implement a subject mapping module to perform structural transformation on the input image according to the structure defined by the extracted landmark heatmap of the auxiliary image. (3) The transformed image is then sent into a second transformation guided by the landmark heatmap of a paired image of the input subject and the final generated image is output. In this manner, we adopt a cycle-like design to complete the transformation cycle between the paired intra-subject images in both directions.

By modeling an intermediate landmark based inter-subject transformation, the landmark detector is enforced to extract semantically-consistent facial landmark locations across different subjects to produce accurate landmark based image generation. The cycle-like intra-subject translation enables additional supervision that encourages our network to learn consistent referential keypoints for both forward and backward image translations. These two factors together help our network to not only extract discriminative landmark locations for each subject in accordance with the provided transformation, but also simultaneously retain landmark semantics across different subjects. 

In summary, our main contributions are as follows: 
%To the best of our knowledge, 
% \vspace{-0.08in}
\begin{itemize}
  \item We propose an unsupervised learning method for image landmark discovery by focusing on both inter and intra landmark consistencies. 
  \item We construct the inter-subject consistency directly through landmark representations with the use of auxiliary images.
%   \vspace{-0.05in}
  \item We model the intra-subject transformation as a cycle and build a two-path end-to-end trainable structure to improve the intra-subject landmark consistency.
%   \vspace{-0.05in}
  \item Comprehensive quantitative and qualitative evaluations on two public facial image datasets demonstrate that the consistent superior landmark localization performances using our method are observed.
\end{itemize}
%%%%%%%%%%%%%%%%%%%%%%%%%%%%%%

%%%%%%%%%%%%%%%%%%%%%%%%%%%%%%
\begin{figure*}[t!]
	\centering
% 	\vspace{2mm}
	\includegraphics[width=0.85\textwidth]{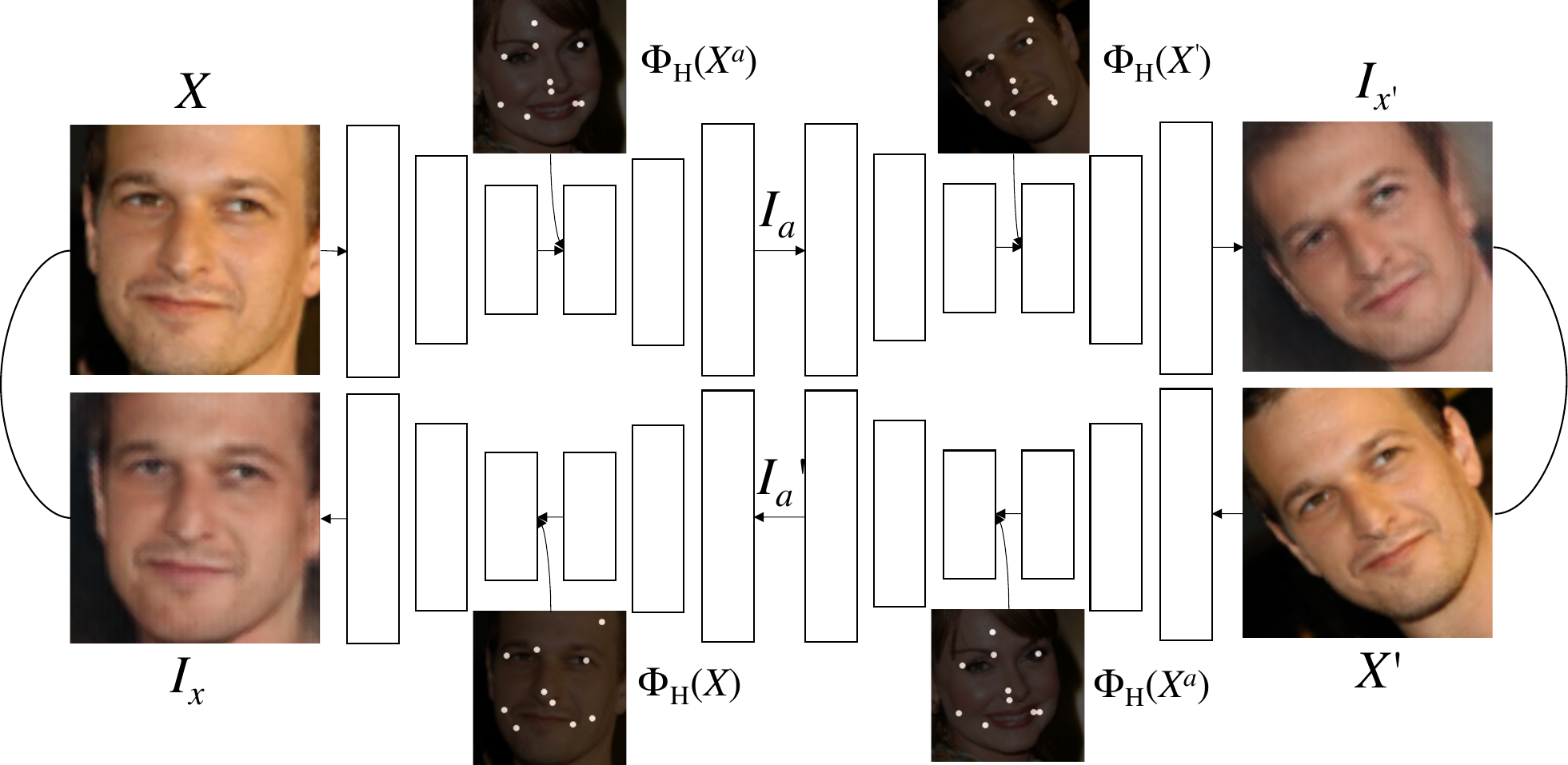}
	\vspace{2mm}
	\caption{Overview of the proposed method. The model takes image $X$ as input and produce the generated image $I_{X'}$ as output. The input image is first transformed by an auxiliary image $X^a$ based on its extracted heatmap $\Phi_H(X^a)$. Then another transformation is applied on the produced image $I_a$ by the paired image $X'$ based on its extracted heatmap $\Phi_H(X')$. A backward cycle path is added at the bottom of the diagram by reversing the top path to complete a cycle-consistency scheme.}
	\label{fig:framework}
	\vspace{-2mm}
\end{figure*}
%%%%%%%%%%%%%%%%%%%%%%%%%%%%%%

%%%%%%%%%%%%%%%%%%%%%%%%%%%%%%
\section{Related Work} 
During recent years, several studies have been conducted towards object landmark discovery with unsupervised learning~\cite{jakab2018unsupervised,sanchez2019object,thewlis2017unsupervised,zhang2018unsupervised,suwajanakorn2018discovery,kulkarni2019unsupervised,wiles2018self,shu2018deforming}. The equivariance~\cite{lenc2016learning} constraint is widely adopted as a supervision signal to learn meaningful landmark locations. For example, Thewlis \textit{et al.}~\cite{thewlis2017unsupervised} propose to extract landmarks compatible with the input image deformations by regressing the probabilistic maps; Suwajanakorn \textit{et al.}~\cite{suwajanakorn2018discovery} detect 3D object keypoints by predicting the known rigid transformations between paired input objects; Zhang \textit{et al.}~\cite{zhang2018unsupervised} introduce a generative framework with learnable landmark locations under a set of transformation constraints. Meanwhile, Jakab \textit{et al.}~\cite{jakab2018unsupervised} also adopt image generation as supervision signal to discover landmarks. They propose to construct a heatmap bottleneck for landmarks by applying the Gaussian-like function centered on the highest responses on the extracted feature map. Then the built Gaussian-like heatmaps act as driving signal to deform input image to the target image. Sanchez \textit{et al.}~\cite{sanchez2019object} study the effect of domain adaptation in unsupervised landmark detection as well as measurement of detection stability by introducing a new evaluation metric. 

While these methods have made great success in different perspectives, inter-subject supervision is usually missing. The paired deformed images are able to help the network locate geometry positions from the same subject but may fail to ensure the position consistency across different subjects. Zhang \textit{et al.}~\cite{zhang2018unsupervised} recognize this issue as \textit{Cross-object correspondence} but they rely on network's implicit learning without supervision. Lately, Thewlis \textit{et al.}~\cite{thewlis2019unsupervised} address this issue by proposing a vector exchange process. During this process, each original image's features at each pixel is first replaced by a weighted aggregation of all pixel features from the auxiliary images. Then maximizing the pixel-wise feature similarities over spatial locations between the exchanged image and the deformed paired image is used as supervision for the learning process. Though our method shares the same idea in the sense to include auxiliary subject images to enable inter-subject information exchange, Thewlis~\cite{thewlis2019unsupervised} \textit{et al.} mainly focus on learning general feature representations and extract landmarks as a separate follow up step. Instead, we directly encode auxiliary images into landmark representations in the same form as the source subject images, and include them as a driving signal to enforce consistent position exchanges that valid for both intra and inter subject relationships.
%%%%%%%%%%%%%%%%%%%%%%%%%%%%%%

%%%%%%%%%%%%%%%%%%%%%%%%%%%%%%
\section{Method} Given an image $x$ $\in\mathcal{X}\subseteq \mathbb{R}^{H \times W \times C}$ and its deformed version $x^\prime \in\mathcal{X}$, our goal is to learn a function $\Phi(x)=y\in\mathcal{Y}$ that extract $K$ structural representations as landmarks without any annotations. Following previous works~\cite{jakab2018unsupervised,sanchez2019object}, we address this problem through conditional image generation. The overall framework can be seen in Figure~\ref{fig:framework} which mainly contains three parts: 1) landmark detectors, 2) inter and intra image generators, and 3) a backward cycle path. This design aims to explore the landmark consistency across the inter- and intra-subject image pairs generated by geometry transformations. In the following sections, we will describe the proposed method in detail.

\subsection{Landmark Detector}
A landmark detector takes an image $x$ as input and outputs $K$ sets of landmark representations where each corresponding to a landmark location. We adopt a similar structure as proposed in~\cite{jakab2018unsupervised,sanchez2019object}. In particular, the input image $x$ is first encoded by a standard convolutional neural network to extract visual feature maps $S\in\mathbb{R}^{H \times W \times K}$. Spatial coordinates for the $K$ landmarks are then obtained from the feature maps and remain differentiable with a Softargmax~\cite{yi2016lift} operation. Specifically, the predicted $k$-th landmark location $u_k$ is the weighted average of the spatial locations $i$, where the weights are computed by the softmax of the $k$-th feature map $S_k$, i.e.,
%%%%%%%%%%%%%%%%%%%%%%%%%%%%%%
\begin{equation}
	u_k = \frac{\sum_{i}exp(\beta S_k(i))i}{\sum_{i}exp(\beta S_k(i))},
	\label{eq:softargmax}
\end{equation}
%%%%%%%%%%%%%%%%%%%%%%%%%%%%%%
where $\beta$ is a hyperparameter for the smoothness. Each prediction is then mapped back to a Gaussian-like probabilistic heatmap centered at $u_k$:
%%%%%%%%%%%%%%%%%%%%%%%%%%%%%%
\begin{equation}
	\Phi_H(x;k) = exp(-\frac{1}{2\sigma^2} \left\|u-u_k \right\|^2 ).
	\label{eq:heatmap}
\end{equation}
%%%%%%%%%%%%%%%%%%%%%%%%%%%%%%
$u_k$ will be the final landmark detection results and $\Phi_H(x;k)$ will be used by later modules as a driving signal to complete the image generation task achieving a self-supervision for $u_k$ learning.

\subsection{Inter-Intra Image Generator}
Previous studies have shown success in unsupervised learning of landmark locations given pairs of images with different geometries. However, since both the two images of a pair come from the same subject, the method does not consider the inter-subject landmark consistency and fails to learn the inter-subject semantics. To this end, we propose to include an auxiliary image which comes from a different subject, and incorporate it as an intermediate transformation as shown in Figure~\ref{fig:framework}. 

Specifically, we denote the geometrically deformed image pairs as $x$ and $x\prime$, and an auxiliary image as $x^a$. An image encoder $\Phi_E$ is first applied to the source image $x$ to extract a visual feature map $\mathcal{F}_s=\Phi_E(x)\in\mathbb{R}^{H \times W \times D}$. At the first stage, we transform the object structure from source image $x$ into auxiliary image $x^a$ based on the landmark representation $\Phi_H(x^a)$ of the auxiliary image, where $\Phi_H$ is the landmark detector we described in the previous section. This is achieved via an image generation function $\Psi$ which takes the concatenation of the visual feature map and the landmark heatmap as inputs, and outputs the generated image:
%%%%%%%%%%%%%%%%%%%%%%%%%%%%%%
\begin{equation}
	\mathcal{I}_a=\Psi(\mathcal{F}_s, \Phi_H(x^a))=\Psi(\Phi_E(x), \Phi_H(x^a)).
\end{equation}
%%%%%%%%%%%%%%%%%%%%%%%%%%%%%%
Next, in the second stage, image $\mathcal{I}_a$ is further transformed by the landmark representations $\Phi_H(x\prime)$ extracted from the paired image $x\prime$. Similarly, we obtain feature map $\mathcal{F}_t$ and the generated target image:
%%%%%%%%%%%%%%%%%%%%%%%%%%%%%%
\begin{equation}
	\mathcal{I}=\Psi(\mathcal{F}_t, \Phi_H(x\prime))=\Psi(\Phi_E(\mathcal{I}_a), \Phi_H(x\prime)).
\end{equation}
%%%%%%%%%%%%%%%%%%%%%%%%%%%%%%
Notice that all three sub-networks are kept the same for both the first and the second stages. In this way, we intentionally inject a dependency of the target image generation on the results of the auxiliary landmark detection which is not available in previous works. In contrast to Thewlis \textit{et al.}~\cite{thewlis2019unsupervised}, our work directly aggregates the landmark detection process on auxiliary images into the model and is more task oriented with end-to-end training. Even with different subject combinations, the entire model is forced to learn only a single set of landmark representations, while at the same time being stable enough to reconstruct any target image. Therefore, each extracted landmark is learned to be consistent on all subject instances.                                                      
\subsection{Cycle Backward Path}
We notice that previous works normally consider the original image $x$ as source image, the deformed image $x\prime$ as the target image to be generated. Similar to Zhu \textit{et al}~\cite{zhu2017unpaired}, we also consider a reversed-order scenario where $x\prime$ is used as the source image and our goal is to reconstruct $x$. The difference is that we focus on learning the landmarks (the conditions) that lead to the generation instead of focusing on the generated results themselves. To achieve this, real facial images $X$ and $X'$ are provided as targets to guide the landmark learning. One may argue that this modification is trivial and can be removed as more deformed image pairs are generated. However, as long as we construct deformed images from $x$ to $x\prime$ by applying a geometrical transformation on $x$, there is always a missing opportunity of  supervision by constructing training target images $x$ transformed from $x\prime$. To complete this, we adopt the same aforementioned network structure, but add a backward cycle path where we switch the source image and target image to $x\prime$ and $x$, as shown in the bottom part of Figure~\ref{fig:framework}. 

\subsection{Training}
Our goal is to learn geometrically meaningful landmarks. This learning process is supervised by accurately generating a deformed image which is driven by these landmarks. To achieve this, we adopt two kinds of losses:

1) reconstruction loss: an MSE loss on the corresponding pixels of the generated image and its groundtruth image which focuses on generation details:
%%%%%%%%%%%%%%%%%%%%%%%%%%%%%%
\begin{equation}
	\mathcal{L}_R(\mathcal{I},\mathcal{I}_{gt})=\left\|\mathcal{I}-\mathcal{I}_{gt}\right\|^2.
\end{equation}
%%%%%%%%%%%%%%%%%%%%%%%%%%%%%%

2) perceptual loss~\cite{johnson2016perceptual}: an MSE loss on the layer outputs of an ImageNet~\cite{deng2009imagenet} pretrained VGG-16~\cite{simonyan2014very} network with the generated target image and its groundtruth image as inputs respectively. It focuses on high level feature representations:
%%%%%%%%%%%%%%%%%%%%%%%%%%%%%%
\begin{equation}
	\mathcal{L}_P(\mathcal{I},\mathcal{I}_{gt})=\sum_{l}\left\|VGG^l(\mathcal{I})-VGG^l(\mathcal{I}_{gt})\right\|^2.  
	   \label{eq:perceptual}
\end{equation}
%%%%%%%%%%%%%%%%%%%%%%%%%%%%%%

% 3) translation loss: the two generated intermediate images should be similar as they are both from the same subject and transformed based on the same auxiliary image. To ensure the effect of intermediate translation, we propose a translation loss $\mathcal{L}_T$ with the same format as the aforementioned $\mathcal{L}_P$ but is applied on $\mathcal{I}_a$ and $\mathcal{I}_a\prime$.

The overall loss is thus a combination of these two losses on both directions of the cycle:
\begin{equation}
	\mathcal{L}=
	\mathcal{L}_R(\mathcal{I}_x, x)
	+\mathcal{L}_R(\mathcal{I}_{x\prime}, x\prime)
	+\mathcal{L}_P(\mathcal{I}_x, x)
	+\mathcal{L}_P(\mathcal{I}_{x\prime}, x\prime)
% 	+\mathcal{L}_T(\mathcal{I}_a, \mathcal{I}_a\prime)
\end{equation}
%%%%%%%%%%%%%%%%%%%%%%%%%%%%%%
Our model is trained end-to-end with the overall loss $\mathcal{L}$.

%%%%%%%%%%%%%%%%%%%%%%%%%%%%%%%%%
\begin{table}[!t]
%\small
\centering
%   \vspace{2mm}
    \caption{Normalized MSE evaluations on the public MAFL and AFLW dataset. Baseline*: our re-implementation of \cite{sanchez2019object}.}
  %\begin{adjustbox}{max width=\columnwidth}
    \resizebox{0.9\linewidth}{!}{
      \begin{tabular}{cccc}
    %   \begin{tabular}{c|c|c|c|c|c|c|c|c|c|c|c}
        \toprule 
            Method & K & MAFL & AFLW \\
            \cmidrule(lr){1-4}
            \multicolumn{4}{c}{Supervised} \\
            TCDCN~\cite{zhang2015learning} & & 7.95 & 7.65 \\
            RAR~\cite{sun2013deep} & & - & 7.23 \\
            MTCNN~\cite{zhang2014facial} & & 5.39 & 6.90 \\
            \cmidrule(lr){1-4}
            \multicolumn{4}{c}{Unsupervised} \\
            Thewlis~\cite{thewlis2017unsupervised2} & - & 5.83 & 8.80 \\
            Shu~\cite{shu2018deforming} & - & 5.45 & - \\
            Sahasrabudhe~\cite{sahasrabudhe2019lifting} & - & 6.01 & - \\
            Wiles~\cite{wiles2018self} & - & 3.44 & - \\
            Thewlis~\cite{thewlis2017unsupervised} & 10 & 7.95 & - \\
            Sanchez~\cite{sanchez2019object} & 10 & 3.99 & 6.69 \\
            Zhang~\cite{zhang2018unsupervised} & 10 & 3.46 & 7.01 \\
            Jakab~\cite{jakab2018unsupervised} & 10 & 3.19 & 6.86 \\
            Zhang~\cite{zhang2018unsupervised} & 30 & 3.15 & 6.58 \\
            Jakab~\cite{jakab2018unsupervised} & 30 & 2.58 & 6.31 \\
            Thewlis~\cite{thewlis2019unsupervised} & 50 & 2.86 & 6.54 \\
            Jakab~\cite{jakab2018unsupervised} & 50 & \textbf{2.54} & 6.33 \\
            \cmidrule(lr){1-4}
            Baseline* & 10 & 3.41 & 6.59 \\
            w. Inter-Subject & 10 & 3.10 & 6.24  \\
            w. Cycle & 10 & 3.12 & 6.28 \\
            Ours-All & 10 & 3.08 & 6.20 \\
            Ours-All & 30 & 2.89 & 6.08 \\
            Ours-All & 50 & 2.85 & \textbf{6.04} \\
        \bottomrule
      \end{tabular}
  }
  %\end{adjustbox}

\label{tab:quant_res1}
\vspace{-2mm}
\end{table}
%%%%%%%%%%%%%%%%%%%%%%%%%%%%%%%%%

%%%%%%%%%%%%%%%%%%%%%%%%%%%%%%
\section{Experiments}
\subsection{Implementation Details} 
\textbf{Landmark Detector}: We follow the previous work~\cite{sanchez2019object} to adopt a Hourglass~\cite{newell2016stacked} based network as our landmark detector which is experimented to be effective in keypoint localization tasks such as Human Pose Estimation, Facial Landmark Detection, etc. It takes $\mathbb{R}^{128 \times 128 \times 3}$ RGB images as input and outputs $\mathbb{R}^{32\times32\times K}$ feature maps. Each heatmap is then transformed to be $\mathbb{R}^{K\times2}$ landmark coordinates $u$ with the Softargmax operation on each $k$th channel of the feature maps. The coordinates are further mapped back to a Gaussian-like heatmap using Equation~\ref{eq:heatmap}. To keep a fair comparison with~\cite{sanchez2019object}, the network is first pretrained on a Human Pose Estimation dataset \textit{MPII}~\cite{andriluka20142d} to detect $K=19$ landmarks. Then all the trained network parameters are fixed. A set of linear projection matrices are applied on the weights of the convolutional layers and are trained for the new detection tasks. We also tried training all the parameters from scratch. The details can be found in the \textit{Ablation Studies} section.

\textbf{Inter-Intra Image Generator}: The image encoder $\Phi_E$ takes $\mathbb{R}^{128 \times 128 \times 3}$ RGB images as input and spatially downsampled the image into a $\mathbb{R}^{32 \times 32 \times 256}$ feature map. It is then concatenated with the obtained landmark heatmaps from the landmark detector $\Phi_H$ along the channel dimension. The concatenated result is sent into the generator network $\Psi$ which contains 6 residual blocks and two spatial upsampling blocks to reconstruct the target image.

\textbf{Learning Facial Landmarks}: To examine the effectiveness of the proposed method, we follow previous works~\cite{jakab2018unsupervised, sanchez2019object, thewlis2017unsupervised, thewlis2019unsupervised} to adopt the CelebA~\cite{liu2015deep} dataset which contains over 200k training images from the celebrities faces excluding 1,000 common images from the MAFL~\cite{zhang2014facial} dataset; the AFLW~\cite{koestinger2011annotated} and MAFL~\cite{zhang2014facial} datasets which contains 10,122/2,991 and 18,997/1,000 training/testing images respectively. During training, the network is first trained on CelebA dataset outputing $K=C$ landmarks as intermediate detections where $C$ is set to be $10$, $30$ or $50$. These landmarks are further linearly regressed into $K=5$ landmarks by training a linear regressor on the AFLW and MAFL training set with all the other parameters of the network fixed. The obtained results are considered the final detection for AFLW and MAFL datasets. To generate geometrically deformed image $x\prime$, a combination of scaling, rotation and translation is applied on the original image $x$. The auxiliary image $x^a$ for each deformed pair $x$ and $x\prime$ is randomly selected from the original images $X$. In our implementation, it is randomly drawn from the other $x$ images in the same batch.

We set parameter $\beta$ in Equation~\ref{eq:softargmax} to $10$, parameter $\sigma$ in Equaltion~\ref{eq:heatmap} equal to $0.5$. The VGG-16 layers we use for the perceptual loss of Equation ~\ref{eq:perceptual} are $l = \{relu1\textunderscore2, relu2\textunderscore2, relu3\textunderscore3, relu4\textunderscore3\}$. Batch size is set to $32$. Adam optimizer is adopted with initial learning rate $0.001$ with a $0.1$ decay rate every $30$ epochs. The proposed model is implemented in PyTorch and is experimented on a single NVIDIA Geforce GTX 1080Ti GPU.

%%%%%%%%%%%%%%%%%%%%%%%%%%%%%%%%%
\begin{table}[!t]
%\small
\centering
%   \vspace{2mm}
    \caption{Normalized MSE evaluations on the MAFL test-set for varying number (N) of supervised samples from MAFL training set used for learning the regressor. We use $K=10$ intermediate landmarks.}
  %\begin{adjustbox}{max width=\columnwidth}
    \resizebox{0.9\linewidth}{!}{
      \begin{tabular}{ccccc}
    %   \begin{tabular}{c|c|c|c|c|c|c|c|c|c|c|c}
        \toprule 
            N & Thewlis K=30~\cite{thewlis2017unsupervised} & Sanchez K=10~\cite{sanchez2019object} & Jakab K=30~\cite{jakab2018unsupervised} & Ours K=10   \\
            \cmidrule(lr){1-5}
            1 & 10.82 & 18.70 & 12.89 & \textbf{9.03} \\
            5 & 9.25 & 8.77 & 8.16 & \textbf{7.50} \\
            10 & 8.49 & 7.13 & 7.19 & \textbf{7.09} \\
            100 & - & 4.53 & 4.29 & \textbf{3.71} \\
            500 & - & 4.13 & \textbf{2.83} & 3.23 \\
            1000 & - & 4.16 & \textbf{2.73} & 3.17 \\
            5000 & - & 4.05 & \textbf{2.60} & 3.09 \\
            All & 7.15 & 3.99 & \textbf{2.58} & 3.08 \\
        \bottomrule
      \end{tabular}
  }
  %\end{adjustbox}

\label{tab:quant_res2}
\vspace{-2mm}
\end{table}
%%%%%%%%%%%%%%%%%%%%%%%%%%%%%%%%%

%%%%%%%%%%%%%%%%%%%%%%%%%%%%%%
\begin{figure*}[t!]
	\centering
% 	\vspace{2mm}
	\includegraphics[width=0.83\textwidth]{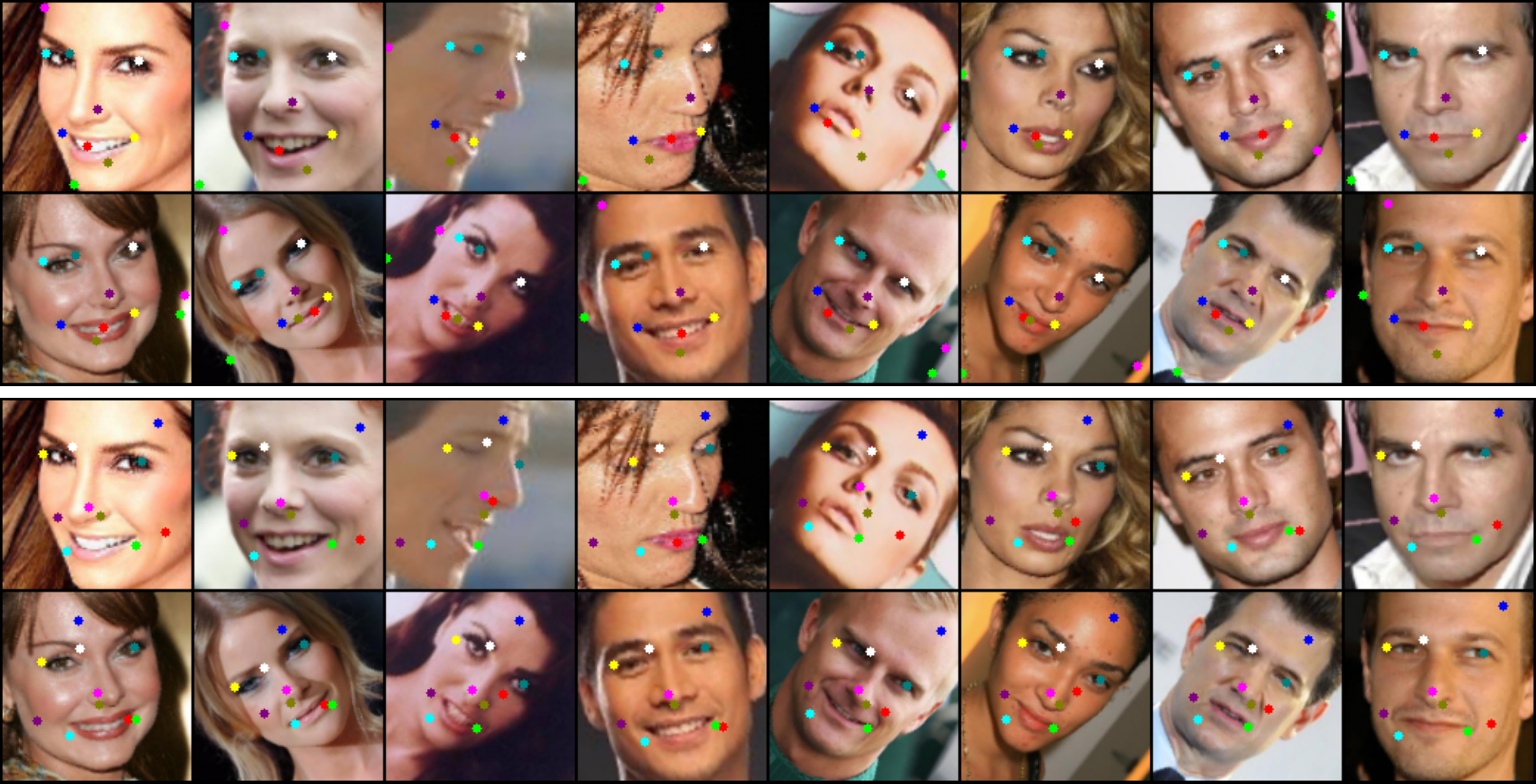}
% 	\vspace{-5mm}
	\caption{Visualization results of the detected landmarks on CelebA. \textbf{Top two rows:} 16 detection results without the proposed modules (Baseline model). \textbf{Bottom two rows:} 16 detection results from the same face subjects as in the top rows but with the proposed modules (Ours full model). Each colored dot corresponding to a detected landmark. Landmarks with the same color within the top or bottom two rows are the same, but are not necessary the same across the top and bottom two rows.}
	\label{fig:visual}
	\vspace{-2mm}
\end{figure*}
%%%%%%%%%%%%%%%%%%%%%%%%%%%%%%

\subsection{Quantitative Evaluation}
Following previous works~\cite{jakab2018unsupervised,thewlis2017unsupervised}, we evaluate the proposed method based on a point-to-point MSE metric normalized by the inter-occular distance on the detected landmarks on the test sets. A baseline method is constructed without the proposed inter-mapping layer and the cycle backward path. As shown in Table~\ref{tab:quant_res1}, integrating the inter-subject mapping module brings improvements comparing to the baseline method indicating the importance of introducing the auxiliary images. Integrating the cycle backward module also improves our model's performance. By combining the two proposed modules together, our complete model \textit{Ours-All} further achieves \textit{3.08\%} and \textit{6.20\%} error rates when detecting $K=10$ intermediate landmarks on the MAFL and AFLW datasets, respectively, which surpass all the strong state-of-the-art unsupervised methods by a large margin indicating the inter-intra compositional effect. Though our method predicts lesser $10$ intermediate landmarks, we notice an even better performance on the AFLW dataset comparing to~\cite{thewlis2019unsupervised} which predicts $50$ intermediate landmarks. For $K=30$ and $K=50$, our model also produce competitive results with a better performance on AFLW when $K$ is set to 50. We believe the reason is that our model is able to locate semantically more meaningful and consistent landmarks for effective inference. This can be further illustrated by examining the visualization results by comparing the detection stability in Figure~\ref{fig:visual}. However, notice that our method perform not as good as Jakab et al.~\cite{jakab2018unsupervised} on the MAFL dataset. We consider a possible reason is the more sophisticated TPS transformation~\cite{duchon1977splines,wahba1990spline} used in~\cite{jakab2018unsupervised} provides enhanced augmented image pairs for the MAFL dataset where images are better aligned than those in the AFLW dataset. This helps for tackling more difficult landmark localization tasks e.g. $K=30$ and $K=50$.

To check our model's capability with different dataset scales, we conduct another evaluation by varying the number of training images when training the final linear regressor on the MAFL dataset following previous works~\cite{thewlis2017unsupervised,sanchez2019object}. As can be seen in Table~\ref{tab:quant_res2}, our method achieves better performance across all the experimental settings comparing to Sanchez et al.~\cite{sanchez2019object} using $K=10$ landmarks for prediction. Notice that a smaller standard deviation is also achieved by our method compared with to~\cite{sanchez2019object}. The results tend to saturate when 100 or more images are used for training, and are almost the same best performance when using 5,000 images and all. It indicates a desirable capability of our method for datasets with less training data. Comparing to Jakab et al.~\cite{jakab2018unsupervised} which use $K=10$ intermediate landmarks our method also achieve competitive results.

\subsection{Qualitative Evaluation}
% To qualitatively examine the detection results and verify the proposed method, we visualize the detected landmarks on the images from CelebA dataset with and without the proposed modules in Figure~\ref{fig:visual}. It is clear to see that most of the landmarks predicted by our model are semantically meaningful, e.g. eye corners, nose, mouse corners, right eye center, etc, while some of the landmarks predicted by the baseline method locate outside facial regions, for example hair tail, neck or collar. We believe these regions should not be considered as valid landmarks since they may not even exist in all the images. Moreover, looking at each detected landmark, we notice that our model is able to extract more consistent locations. For example, comparing the green dots in the top rows and the green dots in the bottom rows, we find both of them tend to focus on the lower face regions. While some of the top green dots drift to other places like chin, background or hair, the bottom row's green dots, on the contrary, are more stable across different face subjects on the lips. However, we also notice some problems predicted by our model. First, although each landmark focus on the same region in general, local variance still exist when occlusion or pose changes occur, such as the pink landmark in the bottom rows seems to prefer the mouse region, but sometime may drift upper or lower marginally. We consider integrating spatial constraint between the landmarks is a possible future direction to explore to achieve better performance.  

%%%%%%%%%%%%%%%%%%%%%%%%%%%%%%%%%
\begin{table}[!t]
%\small
\centering
  \vspace{1mm}
    \caption{Ablation Studies~\cite{sanchez2019object} on the public MAFL and AFLW datasets.}
  %\begin{adjustbox}{max width=\columnwidth}
    \resizebox{0.8\linewidth}{!}{
      \begin{tabular}{cccc}
    %   \begin{tabular}{c|c|c|c|c|c|c|c|c|c|c|c}
        \toprule 
            Method & K & MAFL & AFLW \\
            \cmidrule(lr){1-4}
            VGG-19 & 10 & 3.23 & 6.42 \\
            $\mathcal{L}_R$ only & 10 & 7.87 & 14.98 \\
            $\mathcal{L}_P$ only & 10 & 3.25 & 6.28 \\
            Proposed-All & 10 & 3.08 & 6.20 \\
        \bottomrule
      \end{tabular}
  }
  %\end{adjustbox}

\label{tab:ablative}
\vspace{-6mm}
\end{table}
%%%%%%%%%%%%%%%%%%%%%%%%%%%%%%%%%

To qualitatively examine the detection results and verify the proposed method, we visualize the detected landmarks on the images from CelebA dataset with and without the proposed modules in Figure~\ref{fig:visual}. It is clear to see that most of the landmarks predicted by our method are meaningful, e.g., eye corners, nose, mouse corners, cheek, while some of the landmarks predicted by the baseline method are located outside the facial regions, for example, hair strains, neck or collar. We believe these regions should not be considered as valid landmarks since they may not even exist in all the images. Moreover, looking at each detected landmark, we notice that our model is able to extract more consistent locations. For example, comparing the pink dot in the top rows and the blue dot in the bottom rows, we find both of them tend to focus on the forehead region. While some of the pink dots drift to other places such as the chin, background or hair, the blue dots, on the contrary, are apparently more stable across different face subjects. However, we also notice some problems predicted by our model. Although each landmark focuses on the same region in general, local variance still exists when occlusion or pose changes occur, such as the red landmark in the bottom rows that seems to find the right cheek region, but sometime may drift upper or lower marginally; the blue landmark shifts to the open area when the forehead is covered by hair where the visual appearance is more semantically consistent but geometrically not. Therefore, we consider that integrating landmark spatial constraints will be beneficial for better performance.  

\subsection{Ablation Study} We examine variations of the modules and understand their effects to our model including:1) different network structure for computing Perceptual Loss; 2) different choice of loss function. As shown in Table~\ref{tab:ablative}, the VGG-19 model cannot perform as well as the VGG-16 model. Adopting the reconstruction loss alone is not sufficient to work well for the overall task. When both reconstruction and perceptual losses are adopted, we achieve the best performance. This indicates the importance of the perceptual loss for extracting semantic similarities as well as the benefit from the detailed context information. 

%%%%%%%%%%%%%%%%%%%%%%%%%%%%%%

%%%%%%%%%%%%%%%%%%%%%%%%%%%%%%
\section{Conclusion}
In this work, we introduce an image generation based landmark discovery model with unsupervised learning. Our model extracts inter- and intra-subject consistent landmarks by including (1) an inter-subject mapping module as intermediate translation with auxiliary images from different subjects; (2) a backward cycle path between the original and the translated images from the same subject for additional intra-subject supervision. The superior performance on two public facial image datasets under varies evaluation settings demonstrates the effectiveness of the proposed model.

%%%%%%%%%%%%%%%%%%%%%%%%%%%%%%

%%%%%%%%%%%%%%%%%%%%%%%%%%%%%%
{\small
\bibliographystyle{IEEEtran}
\bibliography{ref}
}
%%%%%%%%%%%%%%%%%%%%%%%%%%%%%%

\end{document}